\documentclass[10pt,twocolumn]{article}

\usepackage[a4paper,margin=0.72in,columnsep=0.24in]{geometry}
\usepackage{fontspec}
\usepackage{amsmath,amssymb}
\usepackage{booktabs,array,tabularx}
\usepackage{graphicx}
\usepackage[font=small,labelfont=bf,skip=4pt]{caption}
\usepackage{subcaption}
\usepackage{xcolor}
\usepackage{tikz}
\usetikzlibrary{arrows.meta,positioning,fit,calc,backgrounds}
\usepackage[numbers,sort&compress]{natbib}
\usepackage{microtype}
\usepackage{xspace}
\usepackage[colorlinks=true,linkcolor=tnblue,citecolor=tnblue,urlcolor=tnblue]{hyperref}
\usepackage{url}

\newfontfamily\khmerfont{NotoSerifKhmer}[
  Path=fonts/, Extension=.ttf,
  UprightFont=*-Regular, BoldFont=*-Bold,
  Script=Khmer, Scale=0.80]
\newcommand{\km}[1]{{\khmerfont #1}}

\newcommand{\sepglyph}{\kern0.04em\rule[-0.14ex]{0.48em}{0.1ex}\kern0.04em}
\catcode"2581=\active
\protected\def▁{\sepglyph}

\definecolor{tnblue}{HTML}{2A78D6}
\definecolor{itnorange}{HTML}{EB6834}
\definecolor{muted}{HTML}{52514E}
\definecolor{surface}{HTML}{F0EFEC}
\definecolor{lightblue}{HTML}{CDE2FB}
\definecolor{lightorange}{HTML}{FBDCCD}

\newcommand{\tha}{\textsc{Tha}\xspace}
\newcommand{\cls}[1]{\texttt{#1}}
\newcommand{\tn}{TN\xspace}
\newcommand{\itn}{ITN\xspace}

\newcommand{\NumPrompts}{2,906}
\newcommand{\NumLexicon}{62,175}
\newcommand{\PyniniVersion}{2.1.7}
\newcommand{\TnStates}{425,482}
\newcommand{\TnArcs}{487,894}
\newcommand{\ItnStates}{246,685}
\newcommand{\ItnArcs}{352,104}
\newcommand{\TnCompile}{4.7}
\newcommand{\ItnCompile}{1.5}
\newcommand{\RtDecimal}{91.0}
\newcommand{\RtTelephone}{98.7}
\newcommand{\RtPerfect}{8}
\newcommand{\RtClasses}{10}
\newcommand{\RtLengthMin}{100.0}
\newcommand{\GoogleTnN}{274}
\newcommand{\GoogleTnExact}{236}
\newcommand{\GoogleTnSpelled}{274}
\newcommand{\GoogleCardN}{265}
\newcommand{\GoogleCardBefore}{228}
\newcommand{\GoogleCardAfter}{265}
\newcommand{\GoogleDecN}{27}
\newcommand{\GoogleDecBefore}{19}
\newcommand{\GoogleDecAfter}{23}
\newcommand{\GoogleTimeN}{13}
\newcommand{\GoogleTimeBefore}{7}
\newcommand{\GoogleTimeAfter}{13}
\newcommand{\PromptChanged}{158}
\newcommand{\PromptErrors}{5}
\newcommand{\PromptCorrect}{153}
\newcommand{\PromptPrecision}{96.8}
\newcommand{\PromptTokens}{172}
\newcommand{\AbFullWords}{36}
\newcommand{\AbFullInside}{4}

\newcommand{\AbFullSent}{158}

\newcommand{\AbNoFilterInside}{9}

\newcommand{\AbSingleWords}{186}
\newcommand{\AbSingleInside}{61}

\newcommand{\AbSingleSent}{438}
\newcommand{\TnMedianMs}{3.4}
\newcommand{\TnPNinetyFiveMs}{6.8}
\newcommand{\TnCharsPerS}{10.9}
\newcommand{\ItnMedianMs}{8.6}
\newcommand{\ItnPNinetyFiveMs}{17.8}
\newcommand{\ItnCharsPerS}{4.4}

\title{\tha: Weighted Finite-State Text Normalization and\\Inverse Text Normalization for Khmer}
\author{Seanghay Yath\\Digital Government Committee, Cambodia\\\texttt{y.seanghay@dgc.gov.kh}}
\date{}

\begin{document}
\maketitle

\begin{abstract}
Text-to-speech needs written text in spoken form, and speech recognition
output needs the reverse. For Khmer, neither direction has a maintained
open-source tool, and the script makes both harder: words are not separated
by spaces, and number words occur inside ordinary words. We present \tha, a
Khmer text normalization and inverse text normalization toolkit built from
weighted finite-state transducers. It segments and classifies a whole line in
one shortest-path search, and a second transducer rejects token boundaries
inside a Khmer syllable. On Google's Khmer test suite, \tha agrees with the
reference on all \GoogleTnN{} cardinals up to one spelling variant, and on
\NumPrompts{} real TTS prompts, \PromptCorrect{} of the \PromptChanged{}
sentences it rewrites are correct. \tha is open source under the Apache 2.0
license.
\end{abstract}

\section{Introduction}
\label{sec:intro}

A TTS front end has to turn ``\km{តម្លៃ} \$1.05'' into the words a speaker
would say, ``\km{តម្លៃ មួយ▁ដុល្លារ▁ប្រាំ▁សេន}'', and an ASR system has to
turn ``\km{ពីររយហាសិបដុល្លារ}'' back into ``\$250''. These two tasks, text
normalization (\tn) and inverse text normalization (\itn), are usually done
with hand-written grammars compiled into weighted finite-state transducers
(WFSTs) \citep{sproat2001normalization,ebden2015kestrel,gorman2016pynini},
since neural models still make the occasional unrecoverable error, such as
reading the wrong number \citep{sproat2016rnn,zhang2019neural}.

For Khmer, the only released grammars are Google's Thrax grammars for TTS
\citep{sodimana2018textnorm}. They do \tn only, expect input that is already
segmented into words, and have been archived since 2021. NeMo
\citep{zhang2021nemo} and WeTextProcessing \citep{wetextprocessing} do not
support Khmer. The language also breaks some of their assumptions: Khmer is
written without spaces between words, so there are no tokens to classify one
at a time, and short number words occur inside ordinary words (\km{ពីរ},
``two'', begins \km{ពីរោះ}, ``melodious'').

\tha is an open-source toolkit for both directions. It tags a whole line in
one shortest-path search, so it needs no word segmenter
(\S\ref{sec:segmentation}), and a filter transducer rejects tokens that start
or end inside a Khmer syllable (\S\ref{sec:filter}). We describe the grammars
for 17 \tn and 13 \itn classes (\S\ref{sec:tn}, \S\ref{sec:itn}) and evaluate
them against Google's test suite, on spoken-form TTS prompts, by round trip
and with an ablation (\S\ref{sec:eval}).

\section{Khmer numbers and script}
\label{sec:khmer}

\paragraph{Script.} Khmer is an abugida of the Brahmic family. A syllable is
a base consonant, optionally stacked with subscript consonants written with
the invisible \emph{coeng} sign (U+17D2), followed by dependent vowel signs
and diacritics \citep{huffman1970cambodian,unicode17}. Unicode allows several
character orders for the same visible syllable, for instance a vowel typed
before a subscript instead of after it, and fonts render them alike; SIL's
Khmer character specification defines a canonical order
\citep{silkhmerspec}. Spaces mark phrase
boundaries rather than words, and the zero-width space (U+200B) is used
inconsistently to mark word boundaries \citep{sodimana2018textnorm}. The
repetition mark \km{ៗ} repeats the preceding word (\km{ក្មេងៗ}, ``children'',
is read \km{ក្មេង▁ក្មេង}), which requires knowing where that word begins.

\paragraph{Number names.} Table~\ref{tab:numbers} lists the Khmer number
words. Six to nine are compounds of five (\km{ប្រាំមួយ}, ``five-one'', is six),
so \km{ប្រាំបួន} can be read as nine or as ``five, four''. The tens have a
spoken short form without \km{សិប} (\km{សាមប្រាំ} for \km{សាមសិបប្រាំ}, 35).
The scale words are \km{រយ} (10\textsuperscript{2}), \km{ពាន់}
(10\textsuperscript{3}), \km{ម៉ឺន} (10\textsuperscript{4}), \km{សែន}
(10\textsuperscript{5}) and \km{លាន} (10\textsuperscript{6}); above a million,
\km{ប៊ីលាន} and \km{ទ្រីលាន} are borrowed, and \km{ពាន់លាន} (``thousand
million'') is common for 10\textsuperscript{9}. Speakers often count
thousands past ten (\km{ដប់ពាន់}, \km{មួយរយពាន់}) instead of using
\km{ម៉ឺន} and \km{សែន}. Several number words are also ordinary words or
names: \km{សែន} means ``very'' and is part of the place name \km{ស្ទឹងសែន},
\km{លាន} is a yard, \km{មួយ} is an indefinite article.

\begin{table}[t]
\centering\small
\setlength\tabcolsep{4pt}
\begin{tabular}{@{}rlrlrl@{}}
\toprule
1 & \km{មួយ} & 10 & \km{ដប់} & 10\textsuperscript{2} & \km{រយ} \\
2 & \km{ពីរ} & 20 & \km{ម្ភៃ} & 10\textsuperscript{3} & \km{ពាន់} \\
3 & \km{បី} & 30 & \km{សាមសិប} (\km{សាម}) & 10\textsuperscript{4} & \km{ម៉ឺន} \\
4 & \km{បួន} & 40 & \km{សែសិប} (\km{សែ}) & 10\textsuperscript{5} & \km{សែន} \\
5 & \km{ប្រាំ} & 50 & \km{ហាសិប} (\km{ហា}) & 10\textsuperscript{6} & \km{លាន} \\
6 & \km{ប្រាំមួយ} & 60 & \km{ហុកសិប} (\km{ហុក}) & 10\textsuperscript{9} & \km{ប៊ីលាន} \\
7 & \km{ប្រាំពីរ} & 70 & \km{ចិតសិប} (\km{ចិត}) & & \km{ពាន់លាន} \\
8 & \km{ប្រាំបី} & 80 & \km{ប៉ែតសិប} (\km{ប៉ែត}) & 10\textsuperscript{12} & \km{ទ្រីលាន} \\
9 & \km{ប្រាំបួន} & 90 & \km{កៅសិប} (\km{កៅ}) & 0 & \km{សូន្យ} \\
\bottomrule
\end{tabular}
\caption{Khmer number words. Short spoken forms of the tens are in
parentheses; they are only used before a digit (\km{ហុកពីរ}, 62).}
\label{tab:numbers}
\end{table}

\paragraph{Written conventions.} Khmer text uses Khmer and Latin digits
interchangeably, often in the same document. Thousands are separated by
spaces, commas or dots (\km{១.០០០} is usually one thousand, not one), decimals
by dots or commas. Dates are written day first. A hyphen between two numbers
is read \km{ដល់} (``to'') in a range, but \km{ទល់} (``against'') in a sports
score. These ambiguities cannot be resolved by a grammar of the token alone,
and \S\ref{sec:tn} describes the context rules \tha applies.

\section{Related work}
\label{sec:related}

\paragraph{Grammar-based normalization.} Text normalization as a task was
laid out by \citet{sproat2001normalization}, with a taxonomy of semiotic
classes later extended by \citet{vanesch2017taxonomy}. Production systems
such as Kestrel \citep{ebden2015kestrel} and its open-source counterpart
Sparrowhawk \citep{sparrowhawk} split normalization into a tokenizer and
classifier, which tags spans with a semiotic class and fields, and a
verbalizer, which turns each tagged token into words. The grammars are
written in Thrax \citep{tai2011thrax} or Pynini \citep{gorman2016pynini} and
compiled into WFSTs with OpenFst \citep{allauzen2007openfst}, using weighted
rewrite rules \citep{mohri1996efficient} and shortest-path search
\citep{mohri1997finite,gorman2021finite}. The same grammars can serve ASR and
TTS \citep{ritchie2019unified}. NeMo \citep{zhang2021nemo} and
WeTextProcessing \citep{wetextprocessing} bring this design to Pynini with
\tn and \itn for many languages; \tha follows NeMo's tagger and verbalizer
interfaces.

\paragraph{Neural and hybrid approaches.} Sequence-to-sequence models
normalize most tokens correctly but make a small number of unacceptable
errors, such as reading the wrong number \citep{sproat2016rnn,sproat2017rnn};
covering grammars \citep{zhang2019neural} and shallow fusion with a WFST
\citep{bakhturina2022shallow} constrain them. Neural \itn has followed the
same path \citep{pusateri2017mostly,sunkara2021neural}. Large language models
have recently been prompted for both \tn \citep{wong2025polynorm,ma2026leveraging}
and spoken-to-written conversion \citep{choi2024spoken}.

\paragraph{Southeast Asian languages.} \citet{sodimana2018textnorm} wrote
Thrax normalization grammars for Bangla, Khmer, Nepali, Javanese, Sinhala and
Sundanese to build TTS voices \citep{sodimana2018stepbystep,openslr42}. Their
Khmer grammars expect input segmented with a CRF word segmenter
\citep{chea2015khmer} and leave segmentation errors to the grammars. Burmese,
also written without spaces, has FST normalization grammars
\citep{oo2020burmese}, and Nisaba provides finite-state script normalization
for Brahmic scripts \citep{johny2021nisaba}.

\paragraph{Khmer.} Khmer is one of the evaluation languages of
\citet{gorman2016minimally}, whose number grammars are induced from a few
examples but cover cardinals only and were not released. \tha uses the
khmercut word segmenter \citep{khmercut} only to expand the repetition mark
(\S\ref{sec:repeat}).
Multilingual speech corpora and models such as FLEURS and Whisper now cover
Khmer \citep{conneau2023fleurs,radford2023whisper}. KhmerTagger \citep{khmertagger} tags number spans and
punctuation in ASR output with XLM-RoBERTa; \tha's \itn grammars could
replace its rule-based converter for the tagged spans (\S\ref{sec:discussion}).

\section{Design}
\label{sec:design}

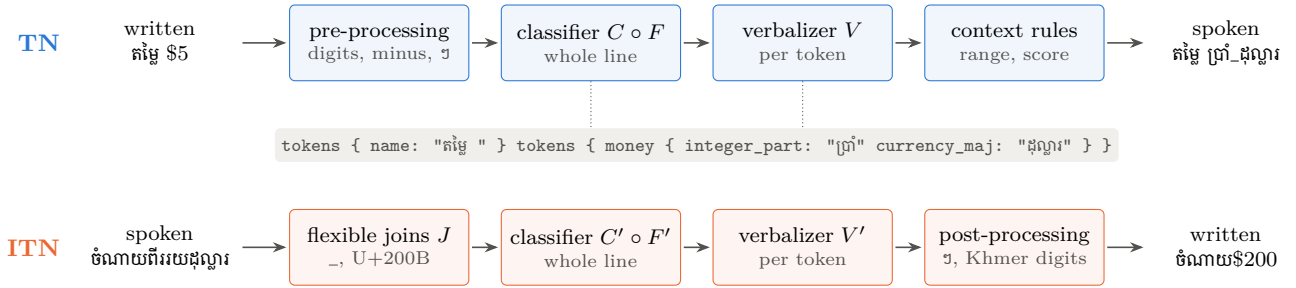
\begin{figure*}[t]
\centering
\begin{tikzpicture}[
  x=1mm, y=1mm,
  box/.style={draw=#1, fill=#1!7, rounded corners=2pt, align=center,
    text width=23mm, minimum height=10mm, inner sep=1pt, font=\footnotesize},
  io/.style={align=center, text width=21mm, inner sep=0pt, font=\footnotesize},
  arr/.style={-{Stealth[length=1.8mm]}, semithick, draw=muted,
    shorten >=1pt, shorten <=1pt},
  row/.style={font=\small\bfseries, text=#1, anchor=east},
]
\def\ya{0} \def\yb{-27}

\node[row=tnblue] at (0, \ya) {TN};
\node[io] (t0) at (12, \ya) {written\\\km{តម្លៃ} \$5};
\node[box=tnblue] (t1) at (41, \ya) {pre-processing\\[-1pt]{\scriptsize\color{muted}digits, minus, \km{ៗ}}};
\node[box=tnblue] (t2) at (69, \ya) {classifier $C \circ F$\\[-1pt]{\scriptsize\color{muted}whole line}};
\node[box=tnblue] (t3) at (97, \ya) {verbalizer $V$\\[-1pt]{\scriptsize\color{muted}per token}};
\node[box=tnblue] (t4) at (125, \ya) {context rules\\[-1pt]{\scriptsize\color{muted}range, score}};
\node[io] (t5) at (153, \ya) {spoken\\\km{តម្លៃ ប្រាំ▁ដុល្លារ}};
\foreach \i/\j in {t0/t1, t1/t2, t2/t3, t3/t4, t4/t5} \draw[arr] (\i) -- (\j);

\node[font=\scriptsize\ttfamily, text=muted, fill=surface, rounded corners=2pt,
      inner sep=2.5pt] (tag) at (83, -13.5)
  {tokens \{ name: "\km{តម្លៃ} " \} tokens \{ money \{ integer\_part: "\km{ប្រាំ}" currency\_maj: "\km{ដុល្លារ}" \} \}};
\draw[muted, densely dotted] (t2.south) -- (t2.south |- tag.north);
\draw[muted, densely dotted] (t3.south) -- (t3.south |- tag.north);

\node[row=itnorange] at (0, \yb) {ITN};
\node[io] (i0) at (12, \yb) {spoken\\\km{ចំណាយពីររយដុល្លារ}};
\node[box=itnorange] (i1) at (41, \yb) {flexible joins $J$\\[-1pt]{\scriptsize\color{muted}\texttt{▁}, U+200B}};
\node[box=itnorange] (i2) at (69, \yb) {classifier $C' \circ F'$\\[-1pt]{\scriptsize\color{muted}whole line}};
\node[box=itnorange] (i3) at (97, \yb) {verbalizer $V'$\\[-1pt]{\scriptsize\color{muted}per token}};
\node[box=itnorange] (i4) at (125, \yb) {post-processing\\[-1pt]{\scriptsize\color{muted}\km{ៗ}, Khmer digits}};
\node[io] (i5) at (153, \yb) {written\\\km{ចំណាយ}\$200};
\foreach \i/\j in {i0/i1, i1/i2, i2/i3, i3/i4, i4/i5} \draw[arr] (\i) -- (\j);
\end{tikzpicture}
\caption{The two pipelines. The classifier tags a whole line in one
shortest-path search (Eq.~\ref{eq:tag}); the tagged form of the \tn example is
shown between the rows. In the \itn direction, each class grammar is composed
with the flexible-join transducer $J$ before it enters the classifier.}
\label{fig:pipeline}
\end{figure*}

Figure~\ref{fig:pipeline} shows both pipelines. Each has a classifier WFST
$C$ that maps a line of text to a tagged token sequence, a filter $F$ that
constrains the tokens, and a verbalizer $V$ that writes each tagged token in
the target form. For input $x$ the tagged form is
\begin{equation}
  t^{*} = \mathrm{ShortestPath}\big(\pi_{\text{out}}(x \circ C) \circ F\big),
  \label{eq:tag}
\end{equation}
where $\circ$ is composition, $\pi_{\text{out}}$ projects onto the output tape
and the shortest path is taken in the tropical semiring. The tagged string
uses NeMo's format,
\begin{center}\small
\texttt{tokens \{ name: "\km{តម្លៃ} " \}}\\
\texttt{tokens \{ money \{ integer\_part: "\km{ប្រាំ}"}\\
\texttt{currency\_maj: "\km{ដុល្លារ}" \} \}}
\end{center}
for ``\km{តម្លៃ} \$5'' (``price \$5''), and is parsed in Python; each semiotic
token is then rewritten by $V$ on its own. Parts of a spoken token are joined
with a separator, \texttt{▁} by default, so that a consumer can tell a
five-word number from five words. The separator can be set to a space or
removed.

\subsection{Segmenting while classifying}
\label{sec:segmentation}

NeMo's classifier consumes one whitespace-separated token at a time. Khmer
has no such tokens, so \tha's classifier covers the whole line. A line is a
sequence of alternating \emph{name} tokens (plain text, passed through) and
semiotic tokens:
\begin{equation}
  C = N^{?}\,(S\,N)^{*}\,S^{?}, \qquad
  S = \textstyle\bigcup_{c} \big(G_c \otimes w_c\big),
\end{equation}
where $G_c$ is the grammar of class $c$ and $w_c$ its weight. A name token $N$
is any non-empty string, and each of its characters costs~1. A semiotic token
costs a flat $w_c < 1$ however long it is (Table~\ref{tab:weights}). The
cheapest path therefore covers as many characters as possible with as few
semiotic tokens as possible: covering five characters with one semiotic token
costs less than one, leaving them as text costs five. Among classes that match
the same span, the smaller $w_c$ wins, which encodes preferences such as
\cls{date} over \cls{cardinal}. Because tokens alternate, two semiotic
tokens are never adjacent, so ``10:234'' cannot be split into a time and a
number.

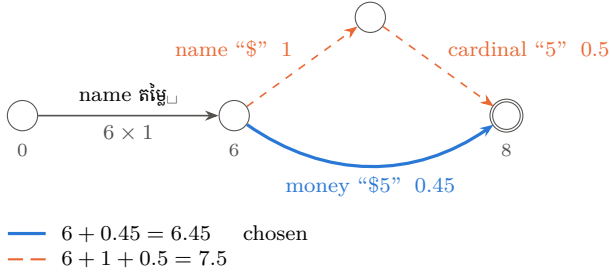
\begin{figure}[t]
\centering
\begin{tikzpicture}[
  x=1mm, y=1mm, font=\footnotesize,
  st/.style={circle, draw=muted, fill=white, minimum size=4mm, inner sep=0pt},
  offset/.style={text=muted, font=\scriptsize},
  arr/.style={-{Stealth[length=1.8mm]}, semithick},
]
\node[st] (s0) at (0, 0) {};
\node[st] (s1) at (28, 0) {};
\node[st] (s2) at (46, 13) {};
\node[st, double] (s3) at (64, 0) {};
\node[offset, below=1pt of s0] {0};
\node[offset, below=1pt of s1] {6};
\node[offset, below=1pt of s3] {8};

\draw[arr, draw=muted] (s0) -- node[above] {name \km{តម្លៃ}\textvisiblespace}
  node[below, text=muted] {$6 \times 1$} (s1);
\draw[arr, draw=itnorange, dashed] (s1) -- node[above left, text=itnorange]
  {name ``\$'' \;$1$} (s2);
\draw[arr, draw=itnorange, dashed] (s2) -- node[above right, text=itnorange]
  {cardinal ``5'' \;$0.5$} (s3);
\draw[arr, draw=tnblue, very thick] (s1) to[bend right=35]
  node[below, text=tnblue] {money ``\$5'' \;$0.45$} (s3);

\node[anchor=north west, inner sep=0pt] at (-2, -14) {%
  \setlength\tabcolsep{3pt}%
  \begin{tabular}{@{}lll@{}}
  \textcolor{tnblue}{\rule[0.5ex]{5mm}{1.2pt}} & $6 + 0.45 = 6.45$ & chosen \\
  \textcolor{itnorange}{\rule[0.5ex]{2mm}{0.8pt}\hspace{1mm}\rule[0.5ex]{2mm}{0.8pt}}
    & $6 + 1 + 0.5 = 7.5$ & \\
  \end{tabular}};
\end{tikzpicture}
\caption{Two paths through the \tn classifier lattice for
``\km{តម្លៃ} \$5''. Numbers under the states are character offsets. Plain
text costs 1 per character and a semiotic token a flat weight, so the path
that covers more text with one token wins. No path can leave the digit 5 as
plain text: digits are not allowed in name tokens.}
\label{fig:lattice}
\end{figure}

Figure~\ref{fig:lattice} shows the competing paths for ``\km{តម្លៃ} \$5''. In
the \tn direction, digits are excluded from name tokens, which guarantees
that every digit is verbalized. Pronounceable symbols (\%, \&, +, \$, °, …)
may appear in name tokens but are replaced by their Khmer names there, so that
nothing unpronounceable reaches a TTS transcript.

\begin{table}[t]
\centering\small
\begin{tabular}{@{}lll@{}}
\toprule
$w_c$ & \tn classes & \itn classes \\
\midrule
0.30 & whitelist & \\
0.40 & date, time, telephone, & date, time, telephone, \\
     & electronic, year range, & electronic \\
     & formation & \\
0.45 & money, measure, ordinal, & money, measure, \\
     & license plate, roman & license plate \\
0.50 & cardinal, fraction & cardinal, fraction, \\
     &                    & range, arithmetic \\
0.55 & decimal, code & decimal \\
0.60 & serial & serial \\
\midrule
1.00 & \multicolumn{2}{l}{per character of a name token} \\
\bottomrule
\end{tabular}
\caption{Class weights $w_c$. A semiotic token costs $w_c$ regardless of its
length; plain text costs 1 per character.}
\label{tab:weights}
\end{table}

\subsection{Boundary filter}
\label{sec:filter}

The cheapest path may still start or end a semiotic token in the middle of a
syllable, for instance by reading the \km{ពីរ} of \km{ពីរោះ} as a number.
The filter $F$ in Eq.~\ref{eq:tag} is an acceptor over tagged strings,
\begin{equation}
  F = \Sigma^{*} \setminus \Sigma^{*}\,B\,\Sigma^{*},
\end{equation}
where $B$ is a union of forbidden patterns at the junction of a name token
and a semiotic token. In the \itn direction these are: a semiotic token
followed by a Khmer dependent vowel, diacritic or \km{ៗ} (which cannot begin a
syllable), a semiotic token preceded by a coeng (a subscript consonant belongs
to the previous one), and a semiotic token whose written form would be glued
to a digit or Latin letter of the neighbouring text (``12\km{ប្រាំ}'' would
otherwise become 125). In the \tn direction the filter keeps ordinals, times,
dates and similar classes away from Latin letters (``1stly'', ``5 mango'' is
not 5\,m + ``ango''), and rejects the dot thousands reading of a number
followed by a scale word (\km{២.១៦៦លាន} is 2.166 million, not 2,166 million).
Applying the constraints to the tagged lattice, rather than inside each
grammar, keeps the class grammars independent of their context.

\subsection{Normalization grammars}
\label{sec:tn}

\paragraph{Cardinals.} Let $\mathit{nz}_k$ read $k$-digit strings without a
leading zero. For $k \ge 3$, with $e$ the largest magnitude in
$\{2,3,4,5,6,9,12\}$ below $k$ and $m_e$ its scale word,
\begin{equation}
  \mathit{nz}_k = \mathit{nz}_{k-e}\; m_e \;
  \big( \mathrm{del}(0^{e}) \;\cup\; \texttt{▁}\,\mathit{full}_e \big),
\end{equation}
where $\mathit{full}_e$ reads any non-zero $e$-digit string. This yields
\km{មួយពាន់▁ពីររយ▁សាមសិបបួន} for 1234 and \km{ដប់លាន} for
10\textsuperscript{7}. Numbers of more than 15 digits and numbers with
leading zeros are read digit by digit. Grouped numbers (1,000,000,
1~000~000, 1.000.000) are accepted, as are number signs (\#5, №5, No.~5 →
\km{លេខ▁ប្រាំ}).

\paragraph{Other classes.} \tha covers 17 classes: cardinal, decimal (dot or
comma, read \km{ចុច} or \km{ក្បៀស}), ordinal, fraction (including vulgar
fractions such as ½), money (symbol or ISO code before or after the amount,
with minor units and scale words), measure (units and percentages), time
(24-hour, 12-hour and the French-style 8h30), date (day first, month first
only when unambiguous), Cambodian telephone numbers, e-mail addresses and
URLs, Cambodian license plates, codes of Latin letters and digits (U19,
MH370), Roman numerals after words that take them (\km{ជំពូក} II), serials
(versions, IP addresses), year ranges and seasons (2022/23), football
formations (4-4-2) and a whitelist (\km{ព.ស.}, COVID-19). The word lists
(units, currencies, months, letter names, …) are TSV files that native
speakers can edit without touching the grammars.

\paragraph{Character order.} The grammars match exact character sequences,
so a number word typed in a non-canonical order would not be recognized.
\tha therefore ships a text cleanup step adapted from SIL's \texttt{khnormal}
\citep{silkhmerspec}, meant to run before either direction: it sorts the
characters of each syllable into canonical order, merges confusable vowel
sequences and removes repeated invisible characters. Unlike the current
upstream version, it keeps coeng \km{ដ} and coeng \km{ត} apart, since they are
pronounced differently. Google's TTS prompts are already in canonical order
(the cleanup changes none of them), so the evaluation in \S\ref{sec:eval}
does not depend on it.

\paragraph{Context rules.} Some readings depend on more than the token. A
pre-processing rewrite rule turns a hyphen into a minus sign only when it is
not attached to a word or number on its left (``$-5$'' but not ``2-3'' or
``COVID-19''). A dash between two quantities is read as a range (\km{ដល់}), or
as a score (\km{ទល់}) when the line contains a sports keyword or team names
on both sides, and not after words such as \km{ថ្ងៃទី} (``the day'').
\km{ម៉ោង} (``hour'') and \km{ថ្ងៃទី} before a time or date are absorbed into
the token so that they are not spoken twice. Dot thousands
(\km{១.០០០} = 1000) and score readings can be switched off.

\subsection{The repetition mark}
\label{sec:repeat}

The mark \km{ៗ} tells the reader to say the preceding word again:
\km{ក្មេងៗ} (``children'') is read \km{ក្មេង▁ក្មេង}. It does not say how much
text the preceding word is. With no spaces between words, the scope has to be
recovered, and it is not always one word: \km{បន្តិចម្ដងៗ} (``little by
little'') repeats the two words \km{បន្តិចម្ដង}, and \km{ពីរថ្ងៃម្ដងៗ}
(``every two days'') repeats three.

\tha segments up to 120 characters before the mark with khmercut
\citep{khmercut}, on the same line and after the last space, because a space
ends the phrase: in \km{រៀងរាល់ថ្ងៃ ម្ដងៗ} only \km{ម្ដង} is repeated. Of the
last three words $w_1 w_2 w_3$, the rules in Table~\ref{tab:repeat} choose
how many to repeat; by default it is the last one. The rules accept both
spellings of \km{ម្ដង}, with coeng \km{ដ} or coeng \km{ត}. A space before the
mark is ignored (\km{ក្មេង ៗ}), and a mark with no word before it is dropped.
The \itn post-processing reverses the expansion: a word or phrase repeated
with \texttt{▁} gets its mark back.

\begin{table}[t]
\centering\small
\setlength\tabcolsep{3pt}
\begin{tabularx}{\columnwidth}{@{}>{\raggedright\arraybackslash}X l l@{}}
\toprule
Last words & Repeats & Example \\
\midrule
any & $w_3$ & \km{ផ្ទះធំៗ} → \km{ផ្ទះធំ▁ធំ} \\
$w_2$ is \km{មួយ}, \km{ទាំង} or \km{លើក} & $w_2 w_3$ & \km{មួយចានៗ} → \km{មួយចាន▁មួយចាន} \\
$w_3$ is \km{ម្ដង}, \km{ឡើង}, \km{ទៅ} or \km{ទៀត} & $w_2 w_3$ & \km{បន្តិចម្ដងៗ} → \km{បន្តិចម្ដង▁បន្តិចម្ដង} \\
\quad and $w_2$ a unit of time & $w_1 w_2 w_3$ & \km{ពីរថ្ងៃម្ដងៗ} → \km{ពីរថ្ងៃម្ដង▁ពីរថ្ងៃម្ដង} \\
\quad and $w_2$ is \km{ពេល} & $w_3$ & \km{ពេលទៅៗ} → \km{ពេលទៅ▁ទៅ} \\
\km{ម្នាក់} and \km{ម្ដង}, either order & $w_2 w_3$ & \km{ម្នាក់ម្ដងៗ} → \km{ម្នាក់ម្ដង▁ម្នាក់ម្ដង} \\
\bottomrule
\end{tabularx}
\caption{How much text \km{ៗ} repeats, decided on the last three words
$w_1 w_2 w_3$ before it as segmented by khmercut. Units of time are
\km{ម៉ោង}, \km{ថ្ងៃ}, \km{យប់}, \km{សប្ដាហ៍}, \km{អាទិត្យ}, \km{ខែ} and
\km{ឆ្នាំ}.}
\label{tab:repeat}
\end{table}

\subsection{Inverse normalization grammars}
\label{sec:itn}

\begin{figure}[t]
\centering
\begin{tikzpicture}[
  font=\footnotesize,
  blk/.style={draw=white, line width=1pt, fill=#1, minimum height=7mm,
    align=center, inner sep=0pt},
]
\def\u{3.3mm}
\node[blk=itnorange!30, minimum width=3*\u, anchor=west] (t) at (0,0) {\km{ទ្រីលាន}};
\node[blk=itnorange!55, minimum width=3*\u, anchor=west] (b) at (t.east) {\km{ប៊ីលាន}};
\node[blk=itnorange!85, minimum width=3*\u, anchor=west, text=white] (m) at (b.east) {\km{លាន}};
\node[blk=tnblue!70, minimum width=3*\u, anchor=west, text=white] (th) at (m.east) {\km{ពាន់}};
\node[blk=tnblue, minimum width=3*\u, anchor=west, text=white] (h) at (th.east) {\km{រយ}};
\foreach \i in {0,...,14} {
  \node[font=\tiny, text=muted] at ({(\i+0.5)*\u}, -0.55) {\pgfmathparse{int(14-\i)}\pgfmathresult};
}
\node[font=\tiny, text=muted, anchor=east] at (-1mm,-0.55) {10\textsuperscript{k}};
\node[align=left, anchor=north west, font=\footnotesize] at (-0.2,-0.9) {%
  \begin{tabular}{@{}ll@{}}
  \multicolumn{2}{@{}l}{\textbf{Below a million} (blue), any of:}\\
  \km{[X សែន][X ម៉ឺន][X ពាន់]} \emph{hundreds} & \km{មួយម៉ឺនប្រាំពាន់}\\
  \emph{hundreds} \km{ពាន់} \emph{hundreds} & \km{ដប់ពាន់}, \km{មួយរយពាន់}\\
  \emph{tens} \km{ម៉ឺន [X ពាន់]} \emph{hundreds} & \km{ដប់ម៉ឺន}\\[2pt]
  \multicolumn{2}{@{}l}{\textbf{Millions}: up to six digits before \km{លាន}}\\
  \multicolumn{2}{@{}l}{\quad when there is no \km{ប៊ីលាន}, so \km{មួយពាន់លាន} $=10^9$}\\
  \multicolumn{2}{@{}l}{\textbf{Above}: \emph{hundreds} before \km{ប៊ីលាន} and \km{ទ្រីលាន}}\\
  \end{tabular}};
\end{tikzpicture}
\caption{The \itn cardinal grammar writes every number as 15 zero-padded
digits, one block per scale word, then removes the leading zeros. Each block
accepts the standard and the colloquial spoken structures.}
\label{fig:blocks}
\end{figure}
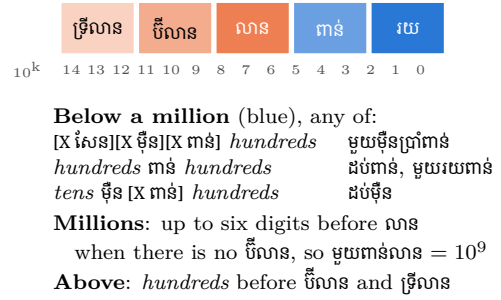

\paragraph{Cardinals.} The \itn cardinal grammar is written directly on the
spoken side rather than by inverting the \tn grammar, so that it accepts forms
the \tn grammar never produces. Following the approach of NeMo's \itn
grammars, it builds a zero-padded 15-digit string, one block per scale word
(Figure~\ref{fig:blocks}), and then deletes the leading zeros. Each block
admits several spoken structures: the six digits below a million can be
\km{[X សែន][X ម៉ឺន][X ពាន់] hundreds}, \km{hundreds ពាន់ hundreds} (the
colloquial \km{ដប់ពាន់}, \km{មួយរយពាន់}) or \km{tens ម៉ឺន …} (\km{ដប់ម៉ឺន}).
Millions are counted up to 999,999, so \km{មួយពាន់លាន} is 10\textsuperscript{9}
without a rule of its own. Short tens (\km{សាមប្រាំ}), the pronunciation
spelling \km{ប្រាំពិល} (7) and zero-padded numbers (\km{សូន្យសូន្យប្រាំពីរ} → 007)
are accepted.

\paragraph{Flexible joins.} Khmer spoken-form text, like any Khmer text,
writes number words together (\km{ពីររយហាសិប}), but the \tn output joins
them with \texttt{▁} and text from the web often has zero-width spaces between
words. Rather than enumerating joiners in every grammar, each class grammar
$G_c$ is written without them and composed with a transducer $J$ that
optionally deletes a \texttt{▁} or U+200B between two characters: $J \circ
G_c$. $J$ is an optional context-dependent rewrite rule
\citep{mohri1996efficient}. A space is not a joiner. It marks a phrase
boundary in Khmer, so \km{ពីររយ ហាសិប} is read as two numbers, 200 and 50;
only the date grammar accepts spaces between its parts, as in \km{ថ្ងៃទីពីរ
ខែមករា}.

\paragraph{Single digits.} A single number word on its own is usually part of
the phrasing (\km{មួយចំនួន}, ``some''; \km{ប្រាំនាក់}, ``five people'') and
is left as a word. It is
written as a digit only inside a larger token (a time, an amount, a
measure, a range) or after \km{ទី} and \km{លេខ} (``\km{ទីបី}'' → ``\km{ទី}3'').

\paragraph{Negative numbers.} \km{ដក} before a number is read as a minus sign
in cardinals, decimals, measures and amounts of money: \km{ដកដប់} → $-10$,
\km{ដកដប់អង្សាសេ} → $-10$\,°C, \km{ដកប្រាំដុល្លារ} → $-\$5$. A single digit
counts here, since the sign makes it a number. Between two numbers \km{ដក}
is subtraction, read by the arithmetic class below. \km{ដក} is also a common verb (``to take
out'', ``to withdraw''), but only a following number turns it into a sign, so
words such as \km{ដកហូត} (``to revoke'') are left alone.

\paragraph{Arithmetic.} The operator words the normalizer produces for
$+$, $-$, $\times$, $\div$ and $=$ (\km{បូក}, \km{ដក}, \km{គុណ}, \km{ចែក},
\km{ស្មើ}) are written back as symbols when they stand between two numbers:
\km{ម្ភៃដកដប់} → 20 - 10, \km{ពីរបូកបីស្មើប្រាំ} → 2 + 3 = 5. Outside that
position they stay words, so \km{ចែករំលែក} (``to share'') and \km{គុណភាព}
(``quality'') are left alone. A single expression token is cheaper than two
numbers around a word, so \km{ម្ភៃដកដប់} is never read as 20 followed by
$-10$.

\paragraph{Classes and output.} The \itn grammars cover 13 classes:
cardinal, decimal, money, measure, time (including \km{កន្លះ}, ``half
past''), full dates, telephone numbers, e-mail addresses and URLs spoken with
``dot''/\km{ចុច} and ``at''/\km{អ៊ែត}, license plates, fractions, ranges and
scores, arithmetic, and serials. The written conventions are options: Latin or Khmer
digits, and a thousands separator for numbers of five digits or more. A word
repeated with \texttt{▁} gets its repetition mark back
(\km{ក្មេង▁ក្មេង} → \km{ក្មេងៗ}, \S\ref{sec:repeat}).

\subsection{Implementation}

\begin{figure}[t]
\centering
\includegraphics[width=\columnwidth]{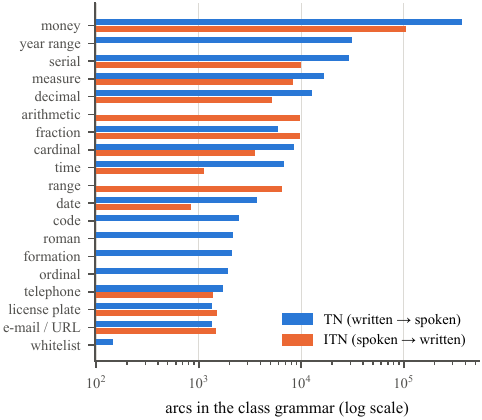}
\caption{Size of each class grammar before it is combined into the
classifier. Money and measure are the largest because every currency and unit
is its own path. Classes without a bar in one direction exist only in the
other.}
\label{fig:sizes}
\end{figure}

\tha is a Python package built on Pynini \PyniniVersion{}
\citep{gorman2016pynini}. The grammars compile when first used, in
\TnCompile{}\,s for \tn and \ItnCompile{}\,s for \itn on a laptop, and are
cached for the process. The compiled \tn classifier has \TnStates{} states and
\TnArcs{} arcs, the \itn classifier \ItnStates{} states and \ItnArcs{} arcs
(Figure~\ref{fig:sizes}). The API mirrors NeMo's:

\begin{center}\small
\begin{tabular}{@{}l@{}}
\texttt{import tha}\\
\texttt{tha.normalize\_text("\km{តម្លៃ} \$1.05")}\\
\quad\texttt{\# \km{តម្លៃ មួយ▁ដុល្លារ▁ប្រាំ▁សេន}}\\
\texttt{tha.inverse\_normalize\_text(}\\
\quad\texttt{"\km{ចំណាយពីររយហាសិបដុល្លារ}")}\\
\quad\texttt{\# \km{ចំណាយ}\$250}\\
\end{tabular}
\end{center}

\section{Evaluation}
\label{sec:eval}

There is no public gold standard for Khmer text normalization, so we use
four smaller evaluations that can all be rerun from public data: agreement
with an independent grammar, \itn on real spoken-form text, round trips, and
an ablation.

\paragraph{Data.} (1) The verbalizer test files of Google's Khmer grammars
\citep{sodimana2018textnorm,googlelanguageresources}: \GoogleTnN{} cardinals,
and decimals and times, each with a structured value and the words the
reference grammar produces. (2) \NumPrompts{} prompts of Google's Khmer TTS
corpus \citep{sodimana2018stepbystep,openslr42}, real sentences already
written in spoken form, with numbers as words. Both sources separate every
word with a space, which Khmer text does not do, so we remove the spaces
between Khmer characters before running \itn: the prompts then read like
ordinary Khmer text, and every number has to be found without word
boundaries. (3) The \NumLexicon{} Khmer
entries of the accompanying pronunciation lexicon, as a list of ordinary words
that \itn should leave alone. (4) Randomly generated written tokens of ten
classes, 300 per class.

\subsection{Agreement with Google's grammar}
\label{sec:google}

Table~\ref{tab:google} compares \tha with the reference verbalizations. In
the \tn direction, \tha's cardinal words equal the reference, spaces
removed, in \GoogleTnExact{} of \GoogleTnN{} cases. All other cases differ
only in the spelling of seven: the reference writes \km{ប្រាំពិល}, following
the pronunciation, where \tha writes the standard \km{ប្រាំពីរ}. With that
spelling mapped, all \GoogleTnSpelled{} agree.

In the \itn direction we read the reference's spoken forms back with \tha;
single digits without a sign, which \tha deliberately leaves as words, are
excluded. Before we added \km{ប្រាំពិល} to the \itn grammar, every
cardinal error came from this spelling. \km{ដប់ប្រាំពិល} (17), for example,
was read as ``15\km{ពិល}'', because the number grammar matched \km{ប្រាំ}
(``five'') and the boundary filter cannot tell that \km{ពិល} continues the
word. With the spelling added, \tha reads all cardinals and times correctly.
The four remaining decimal errors are numbers in scientific notation
(\km{គុណនឹងដប់ស្វ័យគុណ…}, ``times ten to the power of''), which \tha
does not support. We would not have found the \km{ប្រាំពិល} problem with
our own test cases, which all use the standard spelling.

\begin{table}[t]
\centering\small
\setlength\tabcolsep{5pt}
\begin{tabular}{@{}llrrr@{}}
\toprule
& & $n$ & before & after \\
\midrule
\tn & cardinal (exact) & \GoogleTnN & \multicolumn{2}{c}{\GoogleTnExact} \\
    & cardinal (\km{ពិល}~$\equiv$~\km{ពីរ}) & \GoogleTnN & \multicolumn{2}{c}{\GoogleTnSpelled} \\
\midrule
\itn & cardinal & \GoogleCardN & \GoogleCardBefore & \GoogleCardAfter \\
     & decimal  & \GoogleDecN & \GoogleDecBefore & \GoogleDecAfter \\
     & time     & \GoogleTimeN & \GoogleTimeBefore & \GoogleTimeAfter \\
\bottomrule
\end{tabular}
\caption{Agreement with the test suite of Google's Khmer grammars
\citep{sodimana2018textnorm}. \itn columns: correct before and after adding
the spelling \km{ប្រាំពិល} (7) to the \itn grammar.}
\label{tab:google}
\end{table}

\subsection{Inverse normalization of spoken text}
\label{sec:prompts}

\begin{figure}[t]
\centering
\includegraphics[width=\columnwidth]{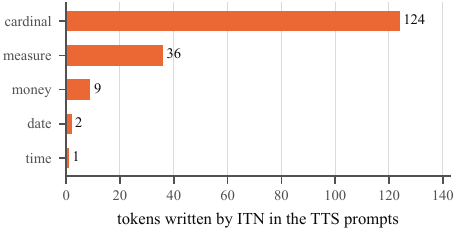}
\caption{Semiotic tokens written by \itn in the \NumPrompts{} TTS prompts.}
\label{fig:classes}
\end{figure}

We ran \itn over all \NumPrompts{} TTS prompts. It rewrote \PromptChanged{}
sentences, writing \PromptTokens{} tokens (Figure~\ref{fig:classes}); the
others contain no number or only single-digit words. We checked every
rewritten sentence by hand. \PromptCorrect{} (\PromptPrecision\%) are
correct, for example \km{តែចិតសិបម៉ែត្រសោះ} → \km{តែ}70\,m\km{សោះ},
\km{ទៅម្ភៃមួយម៉ឺនប្រាំបីពាន់តោន} → \km{ទៅ}218000\,t, the colloquial
\km{ផ្លូវបីរយចិតមួយ} → \km{ផ្លូវ}371 and the full date
\km{ថ្ងៃទីពីរខែមករាឆ្នាំពីរពាន់ដប់បួន} → 02/01/2014. Table~\ref{tab:errors} lists the
\PromptErrors{} errors. Three come from misspelled units in the source text,
one from a unit that is also the beginning of a brand name, and one from
\km{ផោន}, which is both the pound sterling and the pound weight. We also
searched the unchanged sentences for scale and tens words: the only missed
number is the typo \km{សាបសិប} (for \km{សាមសិប}), while \km{ស្ទឹងសែន} (a
river and town), \km{រាប់លាន} (``millions of'') and \km{សែន} as a ritual offering were
correctly left alone. All judgements are the author's.

\begin{table}[t]
\centering\small
\setlength\tabcolsep{3pt}
\begin{tabularx}{\columnwidth}{@{}>{\raggedright\arraybackslash}X>{\raggedright\arraybackslash}p{0.26\columnwidth}@{}}
\toprule
Spoken input (excerpt) & \tha output \\
\midrule
\km{មួយម៉ែត្រការេ} (\km{ការ៉េ} misspelled) & 1 m\km{ការេ} \\
\km{…គីឡូម៉ែតក្រឡា} (\km{ម៉ែត្រ} misspelled) & 574 kg\km{ម៉ែត…} \\
\km{ប្រាំហិកតារ} (\km{ហិកតា} misspelled) & 5 ha\km{រ} \\
\km{ពីរពាន់ដប់ប្រាំវ៉ុលវ៉ូ} (Volvo) & 2015 V\km{វ៉ូ} \\
\km{ទម្ងន់មួយរយផោន} (weight) & \km{ទម្ងន់}£100 \\
\bottomrule
\end{tabularx}
\caption{All \itn errors in the TTS prompts.}
\label{tab:errors}
\end{table}

\subsection{Round trips}
\label{sec:roundtrip}

\begin{figure}[t]
\centering
\includegraphics[width=\columnwidth]{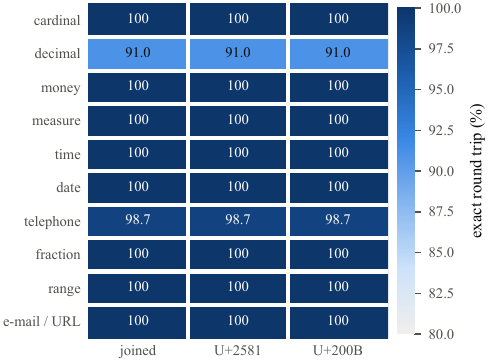}
\caption{Round trip accuracy (\%) by class and by how the spoken words are
joined: glued, with \texttt{▁} (U+2581), or with zero-width spaces (U+200B).
300 tokens per class.}
\label{fig:roundtrip}
\end{figure}

A written token $x$ that \tn verbalizes as $s$ should be written back as $x$
by \itn. We generated 300 written tokens for each of ten classes in \tha's
own written conventions, verbalized them, joined the words of each token in
three ways (glued, \texttt{▁}, zero-width spaces) and read them back
(Figure~\ref{fig:roundtrip}). \RtPerfect{} of the \RtClasses{} classes round
trip exactly in every style. Cardinals of every length from 2 to 15 digits
round trip at \RtLengthMin\% in every style (200 per length).

Decimals round trip in
\RtDecimal\% of cases: the failures are numbers such as 503.811, which \tn,
following Khmer usage, reads as 503811 because a dot is followed by exactly
three digits. The error is in the forward direction and is the documented
cost of the dot-thousands option. Telephone numbers round trip in
\RtTelephone\% of cases because the spoken form itself is ambiguous: the
subscriber number 1030675 is read in chunks as \km{ដប់▁សាមសិប▁ប្រាំមួយរយចិតសិបប្រាំ} (10, 30, 675), which can equally be split as 10, 35,
175. The words alone do not say which one was meant.

\subsection{Ablation}
\label{sec:ablation}

\begin{figure*}[t]
\centering
\includegraphics[width=\textwidth]{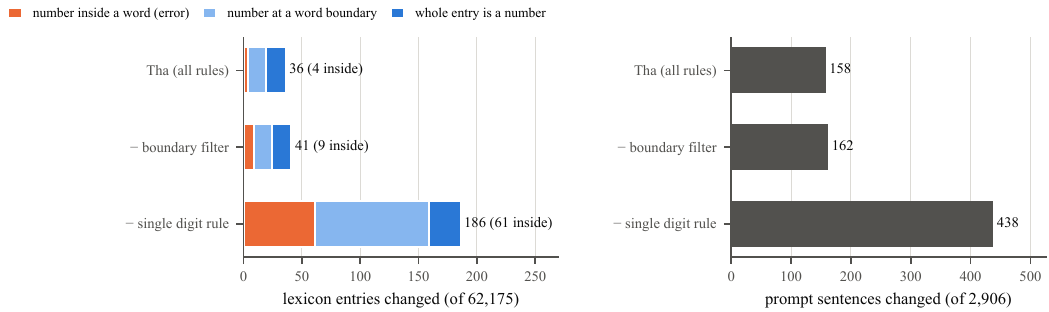}
\caption{Ablation of the \itn design choices. Left: lexicon entries that \itn
changes, by where the number's span falls relative to khmercut's word
boundaries; ``inside a word'' is an error. Right: TTS prompt sentences
changed.}
\label{fig:ablation}
\end{figure*}

We removed one design choice at a time and ran \itn over the lexicon entries
and the prompts (Figure~\ref{fig:ablation}). Most lexicon entries are single
words and should come back unchanged. We classify every change automatically
by aligning the number's span with khmercut's word boundaries: the whole
entry is a number (\km{ដប់}); a number sits at a word boundary of a compound
entry, such as a place name (\km{ភ្នំមួយរយ} → \km{ភ្នំ}100) or an ordinal
(\km{ទីបី} → \km{ទី}3), which is a legitimate reading; or a number starts or
ends \emph{inside} a word, which is an error.

With every rule on, \tha changes \AbFullWords{} of the \NumLexicon{} entries,
and the proxy flags \AbFullInside{} as inside a word. On inspection, three of
these are compounds that khmercut does not split (\km{វិនាទីមួយ}, ``one
second''), and one is a real error: \km{ក្របី} (``buffalo'') ends in
\km{បី} (``three''), a whole syllable that the boundary filter cannot
distinguish from a number, so \km{ក្របីរៀល} becomes \km{ក្រ}3\km{៛}. Without
the boundary filter, \AbNoFilterInside{} entries are split inside a syllable,
for example \km{ដណ្ដប់} (``to cover'') → \km{ដណ្}10, where \km{ដប់} (``ten'')
is a subscript cluster, and \km{មួយរយៈ} → 100\km{ៈ}. Allowing single digits
has by far the largest effect: \AbSingleWords{} entries change,
\AbSingleInside{} of them inside a word (\km{ក្របី} → \km{ក្រ}3,
\km{កំពង់ប្រាំង} → \km{កំពង់}5\km{ង}), and \AbSingleSent{} prompt sentences
change instead of \AbFullSent{}. Some of these extra sentence changes are
correct readings of small numbers (\km{បីខ្នង}, ``three buildings''), so
the rule costs some recall on small numbers. We keep it on by default, since
a missed \km{បី} is harmless and a broken word is not.

\subsection{Speed}

\begin{figure}[t]
\centering
\includegraphics[width=\columnwidth]{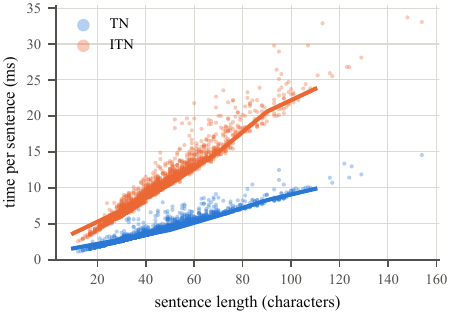}
\caption{Time per sentence against sentence length. Lines are medians over
20-character bins.}
\label{fig:speed}
\end{figure}

Figure~\ref{fig:speed} shows the time per sentence over the prompts, on one
core of a laptop (Apple silicon). The median prompt takes \ItnMedianMs{}\,ms
to inverse normalize and its written form \TnMedianMs{}\,ms to normalize
(95th percentiles \ItnPNinetyFiveMs{} and \TnPNinetyFiveMs{}\,ms), about
\ItnCharsPerS{} and \TnCharsPerS{} thousand characters per second. Time grows
linearly with length, as expected of composition with a fixed transducer.

\section{Discussion and limitations}
\label{sec:discussion}

\paragraph{Ambiguity.} A better grammar will not fix every error. A
telephone number read in chunks, \km{ផោន} as currency or weight, a dash as
range or score, and \km{ដក} before a count (\km{ដកដប់នាក់ចេញ}, ``take ten
people out'', is written $-10$\km{នាក់ចេញ}) all need information that is not
in the token. \tha resolves
the last with sentence-level cues and leaves the first two to its weights; a
language model rescoring the lattice, as in shallow fusion
\citep{bakhturina2022shallow}, or a neural span tagger such as KhmerTagger
\citep{khmertagger} could supply the missing context while \tha's grammars
keep the conversion itself exact.

\paragraph{Orthographic variation.} The Google evaluation and the TTS prompts
both show that the errors that remain come from spellings the grammar does
not know: pronunciation spellings (\km{ប្រាំពិល}), misspelled units and typos.
Because the word lists are data files, these can be added without changing the
grammars, but they must first be found; a list of attested variants from a
large corpus would help most. A related limit of the boundary filter is a
number word that is a whole syllable at the end of another word: \km{បី}
(``three'') ends \km{ក្របី} (``buffalo''), and only a lexicon, not the
script, can tell them apart.

\paragraph{Style choices.} Several outputs are conventions rather than facts:
Latin or Khmer digits, symbols (\km{៛}, \%) or unit words, numeric or spelled
dates, and whether single digits become digits. \tha exposes the most
important as options, but a downstream user may prefer others.

\paragraph{Coverage.} \itn does not read back Roman numerals, letter codes (\km{ជី} is both the
letter G and ``fertilizer'') or whitelist abbreviations. \tn has no
abbreviation expansion beyond a short whitelist.

\paragraph{Evaluation.} Our evaluation uses an independent grammar's test
suite, one corpus of spoken-form text, round trips and an ablation; it does
not include a large, independently annotated gold standard of written Khmer
text, nor a measurement of the repetition rules of \S\ref{sec:repeat}, or a comparison with neural or LLM-based normalizers
\citep{zhang2019neural,wong2025polynorm}. Building such a test set is the
most useful next step.

\section{Conclusion}

\tha brings the tagger and verbalizer design of Sparrowhawk and NeMo to Khmer.
Tagging the whole line at once removes the need for a word segmenter, and the
boundary filter is what allows \itn to run over raw text without breaking
ordinary words. The main open problems are spelling variation and a
proper annotated test set. The code, grammars, experiment scripts and results are available at
\url{https://github.com/seanghay/tha}.

\section*{Acknowledgements}
We thank the authors of Google's Khmer language resources, whose test suite,
TTS prompts and lexicon made the evaluation possible.

\bibliographystyle{plainnat}
\bibliography{refs}

\end{document}